\documentclass[runningheads]{llncs}
\usepackage[utf8]{inputenc}
\usepackage{xspace}
\usepackage{graphicx}
\usepackage{subfigure}
\usepackage{hyperref}
\usepackage{amsmath}

\newcommand\ie{\mbox{i.\,e.}\xspace}
\newcommand\eg{\mbox{e.\,g.}\xspace}
\newcommand\cf{\mbox{cf.}\xspace}

\title{A JIT Compiler for Neural Network Inference}
\author{Felix Thielke \and Arne Hasselbring}
\authorrunning{F. Thielke \and A. Hasselbring}
\institute{Universität Bremen, Fachbereich 3 -- Mathematik und Informatik, \\
Postfach 330 440, 28334 Bremen, Germany \\
\email{\{fthielke,arha\}@uni-bremen.de}}

\begin{document}

\maketitle

\begin{abstract}
  This paper describes a C++ library that compiles neural network models at runtime into machine code that performs inference. This approach in general promises to achieve the best performance possible since it is able to integrate statically known properties of the network directly into the code. In our experiments on the NAO V6 platform, it outperforms existing implementations significantly on small networks, while being inferior on large networks. The library was already part of the B-Human code release 2018~\cite{BHumanCodeRelease2018}, but has been extended since and is now available as a standalone version that can be integrated into any C++14 code base~\cite{URL-Thielke-19}.
\end{abstract}

\section{Introduction}
Within the last years, convolutional neural networks (CNNs) have become a standard solution to image processing and computer vision problems such as object detection and image segmentation. The RoboCup competitions are no exception to this, with many teams using neural networks for the image processing part of their software.
Due to the mostly independent operations within the layers of neural networks, massively parallel GPUs are well suited for their execution. However, they are often not available on mobile robots, due to size, weight or cost reasons. Even for robots that are equipped with a GPU, it might be advantageous to be able to run networks on the CPU, \eg because frames from multiple cameras have to be processed simultaneously or only one large network is offloaded to the GPU while other smaller networks can be run on the CPU to maximize utilization of computational resources.

In order to achieve fast inference of neural networks on CPUs, we developed a C++ library that directly compiles neural network models to optimized machine code at runtime. It is targeted at x86/64 processors with the SSSE3/SSE4 instruction set extensions, especially the Atom and Silvermont microarchitectures present in the NAO and Pepper robots. This is different to most existing libraries that cannot take advantage of static knowledge about the network architecture and are rather optimized for GPUs than those CPUs. By this, we hope to enable the usage of more complex deep learning models on robots not only in RoboCup competitions.

The remainder of this paper is organized as follows: First, Section~\ref{s:related_work} discusses other available libraries for neural network inference. Afterwards, a description of our approach is given in Section~\ref{s:approach}. An evaluation of the performance of our library is presented in Section~\ref{s:evaluation}. Finally, the paper concludes in Section~\ref{s:conclusion}.

\section{Related Work}
\label{s:related_work}

There are several C++ libraries available for inferring neural networks. Of the popular deep learning frameworks, among others, TensorFlow~\cite{tensorflow2015-whitepaper} and Caffe~\cite{jia2014caffe} have C++ interfaces. Both also support training networks in the first place, but are rather heavyweight in their dependencies and mostly optimized for NVIDIA GPUs. TensorFlow includes the TensorFlow Lite library, which is targeted on embedded devices with an ARM processor, but is also able to run on x86 CPUs with SIMD using a NEON/SSE wrapper. Recently, the OpenCV project~\cite{opencv_library} also included support for inferring neural networks via their DNN module. It is able to read file formats of multiple deep learning frameworks and has support for different computation backends, including OpenCL. There are numerous smaller header-only libraries focused on ease of integration and versatility, including tiny-dnn~\cite{tinydnn} and frugally-deep~\cite{frugally_deep}. tiny-dnn also supports training of neural networks, but is not actively developed anymore. frugally-deep is aimed to be compatible to Keras and claims to be faster than Keras itself with the TensorFlow backend on CPUs.

In the context of the RoboCup SPL, members of the Nao-Team HTWK created a fork of the Caffe library, in which they modified the layers for average pooling and the ReLU activation function to use SSE instructions~\cite{HTWKTRR2018}. Szemenyei and Estivill-Castro presented RoboDNN~\cite{Szemenyei2018}, based on the Darknet library, which is specialized on semantic segmentation networks for the NAO.

All of these libraries have in common that they are behaving like interpreters of neural networks, \ie they include branches depending on the actual network structure read from a file that have to be taken on each execution pass. In order to speed up the inference of neural networks, TensorFlow comes with the XLA (Accelerated Linear Algebra) module that just-in-time compiles some layers of a network with optimizations for the target hardware. In addition, a program called tfcompile exists, which uses XLA to compile models ahead-of-time for inference. However, it does not support 32-bit x86 CPUs (thus not being compatible with NAO V5) and the JIT component seems to have a Python interface only.

The Nao Devils Dortmund however used MATLAB to train their neural networks and wrote a script to generate C++ code that performs the inference of the networks on the robot~\cite{NaoDevilsTR2018}. More recently they implemented the same approach in the Python language. This concept resembles our proposed library in principle, except that the compilation of networks takes place ahead-of-time, not at runtime. Apart from that, by generating C++ code, the optimality of their solution depends on the optimizations done by the C++ compiler, while our just-in-time compiler has complete control over the generated code and can thus theoretically produce the most efficient code without depending on external factors. Furthermore, it supports only a very limited selection of layers.

\section{Approach}
\label{s:approach}
In order to achieve the best possible performance, the code for inferring a neural network can be optimized based on several static pieces of information about the model. Thus, we chose to implement our module in such a way that it reads pretrained models at runtime and translates them to machine code exposed as C++ functions by means of the AsmJit library~\cite{URL-Kobalicek-19}.

While this approach allows for every possible optimization of the computations with respect to the target hardware, compiling the code just-in-time also provides flexibility and rapid deployment without an additional export step. However, it slightly increases the startup time of the software. In Sect.~\ref{s:evaluation}, the impact of this is measured for some typical models.

Considering that the main target hardware of our development, the NAO robot, contains an Intel Atom CPU, the code that our module generates is also targeted at this platform. This means that currently only x86 machine code can be emitted, making use of Streaming SIMD Extensions (SSE) up to version SSE4.2 for vectorized operations. Since the NAO's CPU does not have AVX support, we currently do not use it.

\subsection{Front End}
The front end and interface to the programmer of our library is represented by the \texttt{Model} class. Instances of this class contain a neural network architecture---\ie a computational graph of layers---as well as the weights belonging to certain layers of that network. Currently, the \texttt{Model} class allows to load a network only from an HDF5 file as written by the Python library Keras~\cite{chollet2015keras}. For this purpose, our library depends on the HDF5 library~\cite{hdf5} and includes a custom implementation of a JSON parser to obtain the model architecture. The \texttt{Model} class could potentially be extended to support other file formats.

In order to compile the inference code for such a model, it can be passed to an instance of the class \texttt{CompiledNN}. Once the code has been compiled, the method \texttt{apply()} of the \texttt{CompiledNN} instance can be used at any time to call the generated code that calculates the forward pass of the network. The input and output tensors of the network are owned by \texttt{CompiledNN} because it needs control over the actual memory layout.

In addition to this main functionality, the library also includes the class \texttt{SimpleNN}, which provides a straightforward, but slow implementation of neural network inference in C++. As this class was written to be as exact in its calculations as possible, it can be used to benchmark the compiler in terms of numeric precision.

\subsection{Intermediate Processing}
Internally, \texttt{CompiledNN} converts the nodes of the given network to a sequence of compilation units. Generally, each layer is mapped to one compilation unit, however sometimes layers are merged together or split into multiple compilation units if that is deemed beneficial for the compilation process or the performance of the generated code (\cf~Sect.~\ref{s:merging}). Next, the inputs and outputs of all nodes are assigned to actual memory locations, taking into account that tensors with overlapping lifetimes must use different memory. At this stage, the individual layer compilers can indicate whether they want any of their outputs to use the memory of an input tensor that is not referenced afterwards. This way, many compilers can operate in-place, which results in better cache usage and less pointer register operations.

\subsection{Common Principles}
The performance of an x86 instruction depends on its latency and throughput on the given processor architecture. In this context, latency means the number of clock cycles the instruction takes until its result is available, which is relevant in dependency chains. The throughput on the other hand denotes the number of instructions of the same kind that can be executed per clock cycle, given that their operands are independent. Thus, the reciprocal throughput can be read as the number of clock cycles that an instruction takes if it follows a similar instruction and all its operands have already been computed \cite{agner_instruction_tables}.

On the microarchitectures that we target, the latency of all SIMD instructions is either larger than or equal to their reciprocal throughput. Thus, the code generated by our compiler generally follows a certain pattern to make sure that the limit for the performance of the generated code is the throughput of the instructions instead of the latency.
In order to achieve this, our code subdivides the values to be computed into batches of up to \(4\cdot (n_\text{xmm}-k)\) elements, where \(n_\text{xmm}\) is the number of 128-bit XMM registers and \(k\) is an operation specific number of registers that are needed for intermediate results or weights (usually \(2\)). Within each batch, first, all inputs are loaded into the registers. Then the operations, \eg multiplications, are performed on all registers successively. Afterwards, the results are written to the destination addresses.

The core layers of CNNs are convolutional and fully connected layers. While the operation of a fully connected layer can be interpreted as a multiplication of a weight matrix with an input vector, the operation of a convolutional layer consists of a subdivision of the 3D input tensor along the width and height dimensions, followed by a series of multiplications of a kernel matrix with each of the resulting input vectors.
Thus, the matrix-vector-product is the most important operation in our implementation. For this operation the compiler therefore generates specialized versions for several cases concerning the dimensions of the parameters.
Besides, the calculation itself follows a scheme that uses the resources of the processor as well as possible. Using SIMD instructions, up to four floating point numbers can be used in a calculation at a time. The $4 \times 4$ matrix-vector-product as shown in Eq.~\ref{eq:mvp} could then be implemented as shown in Eq.~\ref{eq:mvp_impl1}, where $\odot$ means element-wise multiplication.

The implementation in Eq.~\ref{eq:mvp_impl1} would then need at least three registers: one register in which one column of the matrix is loaded at a time, one containing the input vector $\textbf{x}$ and a third register that successively holds each of the four shuffled versions of the second register, containing only one of its elements at a time. The instructions needed are five load operations for the elements of the matrix and the input vector, four element-wise multiplications and four shuffle operations to broadcast the input vector.

However, the implementation that we use is shown in Eq.~\ref{eq:mvp_impl2}. Note that the elements of the matrix are parameters of the neural network known at compile time, so the memory layout of the matrix can be chosen arbitrarily without any impact on performance. This implementation allows to keep the elements of the input vector in the same register at all times, thus needing one less register. At the same time, it also needs only three shuffle operations as for the first element-wise multiplication, the input vector can be used directly. Saving the one register is crucial in this case as it increases the number of channels that can be computed per batch by \(4\).

\begin{align}
  \label{eq:mvp}
  \begin{pmatrix}y_1 \\ y_2 \\ y_3 \\ y_4\end{pmatrix} &\!=\! \begin{pmatrix}a_{11} & a_{12} & a_{13} & a_{14} \\ a_{21} & a_{22} & a_{23} & a_{24} \\ a_{31} & a_{32} & a_{33} & a_{34} \\ a_{41} & a_{42} & a_{43} & a_{44}\end{pmatrix} \cdot \begin{pmatrix}x_1 \\ x_2 \\ x_3 \\ x_4\end{pmatrix} \\
  \label{eq:mvp_impl1}
  \begin{pmatrix}y_1 \\ y_2 \\ y_3 \\ y_4\end{pmatrix} &\!=\! \begin{pmatrix}a_{11} \\ a_{21} \\ a_{31} \\ a_{41}\end{pmatrix} \!\odot\! \begin{pmatrix}x_1 \\ x_1 \\ x_1 \\ x_1\end{pmatrix} \!+\! \begin{pmatrix}a_{12} \\ a_{22} \\ a_{32} \\ a_{42}\end{pmatrix} \!\odot\! \begin{pmatrix}x_2 \\ x_2 \\ x_2 \\ x_2\end{pmatrix} \!+\! \begin{pmatrix}a_{13} \\ a_{23} \\ a_{33} \\ a_{43}\end{pmatrix} \!\odot\! \begin{pmatrix}x_3 \\ x_3 \\ x_3 \\ x_3\end{pmatrix} \!+\! \begin{pmatrix}a_{14} \\ a_{24} \\ a_{34} \\ a_{44}\end{pmatrix} \!\odot\! \begin{pmatrix}x_4 \\ x_4 \\ x_4 \\ x_4\end{pmatrix} \\
  \label{eq:mvp_impl2}
  \begin{pmatrix}y_1 \\ y_2 \\ y_3 \\ y_4\end{pmatrix} &\!=\! \begin{pmatrix}a_{11} \\ a_{22} \\ a_{33} \\ a_{44}\end{pmatrix} \!\odot\! \begin{pmatrix}x_1 \\ x_2 \\ x_3 \\ x_4\end{pmatrix} \!+\! \begin{pmatrix}a_{12} \\ a_{23} \\ a_{34} \\ a_{41}\end{pmatrix} \!\odot\! \begin{pmatrix}x_2 \\ x_3 \\ x_4 \\ x_1\end{pmatrix} \!+\! \begin{pmatrix}a_{13} \\ a_{24} \\ a_{31} \\ a_{42}\end{pmatrix} \!\odot\! \begin{pmatrix}x_3 \\ x_4 \\ x_1 \\ x_2\end{pmatrix} \!+\! \begin{pmatrix}a_{14} \\ a_{21} \\ a_{32} \\ a_{43}\end{pmatrix} \!\odot\! \begin{pmatrix}x_4 \\ x_1 \\ x_2 \\ x_3\end{pmatrix}
\end{align}

\subsection{Activation Functions}
In neural networks, activation functions are operations that are applied on each input element independently, usually even independent of the position of the element in the input tensor.
Therefore, they can not only be implemented in-place, meaning that the addresses of input and output in memory are the same, but also be appended to other operations, namely fully connected and convolutional layers.
In that case, the activation function is applied before writing the result of the operation into memory. This avoids an additional loop with load and store operations.

However, some activation functions cannot be implemented that way. For example, \emph{Softmax} needs two passes---one to calculate $x'_i = e^{x_i}$ for every input element $x_i$ while at the same time calculating $\sum_i x'_i$ and a second pass to divide all resulting elements $x'_i$ by this sum.
In this case, the activation function is always compiled as a separate compilation unit.

Many usual activation functions utilized by neural networks, among them Softmax and the logistic function, need to calculate the exponential function. This poses a problem for our approach since there is no straightforward way of calculating $\text{exp}(x)$ using SSE instructions.
So instead, we chose to approximate these functions. For instance, the logistic function (sigmoid) can be expressed in terms of the tanh function (see Eq.~\ref{eq:sigmoid}) which in turn can be approximated by calculating a certain amount of steps of the continued fraction that converges to it (see Eq.~\ref{eq:tanhapprox}). This reduces the operations needed to multiplications, additions and one division.

\begin{equation}
  \label{eq:sigmoid}
  \text{sigmoid}(x) = \frac{\text{tanh}\left(\frac{x}{2}\right) + 1}{2}
\end{equation}
\begin{equation}
  \label{eq:tanhapprox}
  \text{tanh}(x) \!=\! \frac{x}{1 + \frac{x^2}{3 + \frac{x^2}{5 + \dots}}} \!\approx\! \frac{(((36 x^2 + 6930) x^2 + 270270) x^2 + 2027025) x}{(((x^2 + 630) x^2 + 51975) x^2 + 945945) x^2 + 2027025}
\end{equation}

Alternatively, the exponential function can easily be approximated by exploiting the characteristics of the IEEE-754 floating point representation. Using the method described in \cite{Schraudolph99}, $\text{exp}(x)$ can be calculated by one multiplication, one float-to-integer conversion and one integer addition, afterwards interpreting the result as a floating point number again. Approximating activation functions however impacts the precision of the calculations, which could lead to different outputs of the neural network.

\subsection{Merging}
\label{s:merging}
Similarly to the aforementioned combination of activation functions into other compilation units, some layers can be eliminated entirely by merging them into adjacent ones. Specifically, batch normalization layers operate by multiplying a vector element-wise with the input tensor along the feature axis and afterwards adding an offset to it. If such a layer is immediately preceded or followed by a convolutional or fully connected layer, these calculations can be eliminated by removing the batch normalization step and adjusting the weights and biases of the other layer in such a way that they already include the factors and offsets of the normalization. Note that this changes the associativity of the calculations, which---because of the limited resolution of the representation of floating point numbers---might lead to a different result compared to the usual execution in two steps. If there is an activation function between the layers, the batch normalization is still fused into the other layer and applied after the activation, since it saves another loop and memory loads and stores.

\section{Evaluation}
\label{s:evaluation}

To evaluate our library quantitatively, we benchmarked it on some models against other available inference libraries. These are frugally-deep~\cite{frugally_deep}, RoboDNN~\cite{Szemenyei2018}, TensorFlow Lite~\cite{tensorflow2015-whitepaper}, and tiny-dnn~\cite{tinydnn}. The selected models include four networks relevant to the RoboCup SPL: The classifier that is used by Nao-Team HTWK~\cite{HTWKTRR2018}, the ball classifier from B-Human~\cite{BHumanCodeRelease2018}, a detection network that predicts bounding boxes of robots on an entire camera image~\cite{Poppinga-RCIS-2019}, and a network suitable to perform semantic segmentation into field/non-field on an 80x80 input. For comparison, MobileNetV2~\cite{mobilenetv2} (\(\alpha=1\), without top) and VGG19~\cite{vgg} have also been evaluated.

All libraries and the evaluation code have been compiled with optimizations enabled and the target architecture set to Silvermont. The runtimes are the average over multiple successive calls to the inference routine, after doing some unmeasured initial runs. Since RoboDNN and tiny-dnn do not support upsampling and depthwise separable convolution layers, the detection and segmentation networks and MobileNetV2 could not be tested. Table~\ref{tab:runtimes} shows the results.

\begin{table}[tb]
  \centering
  \caption{Inference times on the NAO V6's Intel Atom E3845 in milliseconds. C-HTWK denotes the classification network of Nao-Team HTWK, C-BH is the ball classifier from B-Human. The best runtime is marked bold. The last row lists the time in milliseconds our library needs to load and compile each network.}
  \begin{tabular}{l|r r r r r r}
     & C-HTWK & C-BH & Detector & Segmenter & MobileNetV2 & VGG19 \\
    \hline CompiledNN & \textbf{0.007} & \textbf{0.0447} & \textbf{1.995} & \textbf{7.859} & \textbf{145.1} & 14993 \\
    \hline frugally-deep & 0.1724 & 0.5167 & 28.49 & 32.51 & 1036 & 11872 \\
    \hline RoboDNN & 0.0394 & 0.1383 & - & - & - & 20860 \\
    \hline TensorFlow Lite & 0.04276 & 0.3995 & 5.798 & 23.07 & 191.8 & \textbf{10220} \\
    \hline tiny-dnn & 0.1133 & 0.5297 & - & - & - & 100137 \\
    \hline
    \hline Compilation Time & 6.5 & 9.5 & 26.6 & 18.1 & 335 & 13722
  \end{tabular}
  \label{tab:runtimes}
\end{table}

From an application perspective, our library allows the soccer SPL team B-Human to classify many more ball candidate patches per frame than any of the other solutions. This, in combination with a rather sensitive candidate generator, enables us to check for the ball in many locations of the image, decreasing the chance of missing it due to limited computational resources.

\section{Conclusion and Future Work}
\label{s:conclusion}
In this paper, we presented a C++ library that compiles neural network models into machine code that performs inference.
It is targeted at x86/64 processors, especially the Intel Atom and Silvermont microarchitectures that are used in the NAO and Pepper robots. The performance achieved on networks that are generally feasible to execute for each camera frame is significantly better than existing libraries, while being slower for particularly large models. The library is available on GitHub~\cite{URL-Thielke-19}.

We are currently far from supporting all Keras layer types with all possible parameters, although the most commonly used ones are available. Some layers have limitations regarding the input dimensions. Optimizing the overall performance of the library is an ongoing effort without a foreseeable end.

\bibliography{bibliography}

\end{document}